# The CM Algorithm for the Maximum Mutual Information Classifications of Unseen Instances

*Chenguang Lu*

lcguang@foxmail.com

**Abstract**

The Maximum Mutual Information (MMI) criterion is different from the Least Error Rate (LER) criterion. It can reduce failing to report small probability events. This paper introduces the Channels Matching (CM) algorithm for the MMI classifications of unseen instances. It also introduces some semantic information methods, which base the CM algorithm. In the CM algorithm, label learning is to let the semantic channel match the Shannon channel (Matching I) whereas classifying is to let the Shannon channel match the semantic channel (Matching II). We can achieve the MMI classifications by repeating Matching I and II. For low-dimensional feature spaces, we only use parameters to construct *n* likelihood functions for *n* different classes (rather than to construct partitioning boundaries as gradient descent) and expresses the boundaries by numerical values. Without searching in parameter spaces, the computation of the CM algorithm for low-dimensional feature spaces is very simple and fast. Using a two-dimensional example, we test the speed and reliability of the CM algorithm by different initial partitions. For most initial partitions, two iterations can make the mutual information surpass 99% of the convergent MMI. The analysis indicates that for high-dimensional feature spaces, we may combine the CM algorithm with neural networks to improve the MMI classifications for faster and more reliable convergence.

**Keywords**: Maximum mutual information, Maximum likelihood, Unseen instance classifications, Shannon's channel, Semantic channel, Statistical learning

## 1   Introduction

To classify every unseen instance in a true class *X* according to its observed feature *Z* is to provide a classifier *Y*=*f*(*Z*) to get a label *Y* (see Fig. 1). There are various criteria, such as the Maximum Likelihood (ML) criterion, the Least Loss (LL) criterion, the Least Error Rate (LER) criterion, the Regularized Least Square (RLS) criterion, and the Maximum Mutual Information (MMI) criterion. Mutual Information (MI) is defined by Shannon [1]. Since Mutual Information (MI) ascertains the upper limit of average log likelihood [2, 3], this criterion is compatible with the ML criterion.

Shannon and most researchers of the classical information theory use the least loss (or distortion) criterion [1] instead of the MMI criterion to optimize tests, estimations, and unseen instance classifications because it is very difficult to use the MMI criterion. Without the classifier *f*(*Z*), we cannot express mutual information *I*(*X*; *Y*) whereas, without the expression of mutual information, we cannot optimize the classifier *f*(*Z*). Recently the MMI criterion for estimations [4-7], classifications [8-11], and feature selections [12] had shown its advantages and attracted more researchers' attention. Based on the semantic or generalized information theory [13-15], Lu proposed a new MMI classification method, the Channels Matching (CM) algorithm. In his recent paper [3], he introduced the CM algorithm for tests and estimations, including simple classifications for one-dimensional feature spaces, with the MMI criterion. For one-dimensional feature spaces, partitioning boundaries are only one or several points. However, for multi-dimensional feature spaces, partitioning boundaries are lines or curved surfaces, and hence it is very difficult to solve partitioning boundaries. To test the speed, reliability, and generality of the CM algorithm for the MMI classifications, we applied this algorithm to classifications for two-dimensional feature spaces where partitioning boundaries are several curves. The results are inspiring. In this paper, we mainly introduce new experimental results. We also compare the CM algorithm with the popular gradient descent and discusses its applications to high-dimensional feature spaces with neural networks. For completeness of the paper, we also simply introduce the semantic information methods mainly developed by Lu, which base the CM algorithm.

In the next section, we introduce the semantic information methods. In Section 3, we introduce the CM algorithm and a two-dimensional iteration example with different initial partitions. Section 4 provides discussion. The last section includes conclusions.

A video file to show the iteration processes can be obtained from http://survivor99.com/lcg/cm/TestMMI.rar (20M)

## 2   The Semantic Information Methods Basing the CM algorithm

*2.1 Mathematical methods*

Definition 1:

- *x*: an instance or true class (for unseen instance classifications); *X*: a random variable taking a value $x \in U=\{x_0, x_1, …\}$.
- *y*: a hypothesis or label; *Y*: a random variable taking a value $y \in V=\{y_0, y_1, …\}$.

- $z$: an observed feature or feature vector for an unseen instance; $Z$: a random variable taking a value $z \in C=\{z_0, z_1, \ldots\}$. $Z$ is also considered as a factor or factor vector [16].
- $\theta$: a model or a set of model parameters. For given $y_j$, $\theta=\theta_j$ so that $P(x|\theta_j)=P(x|y_j,\theta)$ and $P(\theta_j|x)=P(y_j|x,\theta)$. The $\theta_j$ is also a fuzzy set [17] and $y_j(x)=\text{"}x \in \theta_j\text{"}$.
- $T(\theta_j|x)$ is the truth function of $y_j$ or the membership function of $\theta_j$, which ascertains the semantic meaning of $y_j$ (according to Davidson's truth condition semantics [18]).

According to Tarski's truth theory [19], $P(x\epsilon\theta_j)$ is equivalent to $P(\text{"}x\epsilon\theta\text{"}$ is true$)=P(y_j$ is true$)$. So, $P(x\epsilon\theta_j)$ is the logical probability of $y_j$. It is different from the statistical probability $P(y_j)$ of $y_j$. To distinguish $P(y_j)$ and $P(y_j$ is true) better, we use $T(\theta_j)$ to denote $P(y_j$ is true) or $P(x \in \theta_j)$. There is

$$T(\theta_j) = \sum_i P(x_i)T(\theta_j|x_i) \quad (1)$$

which was proposed by Zadeh as the probability of a fuzzy event [19].

*2.2 The Third kind of Bayes' Theorem*

There are three kinds of Bayes' theorem, which are used by Bayes, Shannon, and Lu respectively [21]. The Third Kind of Bayes' Theorem includes two asymmetrical formulas:

$$P(X|\theta_j) = \frac{T(\theta_j|X)P(X)}{T(\theta_j)}, T(\theta_j) = \sum_i P(x_i)T(\theta_j|x_i) \quad (2)$$

$$T(\theta_j|X) = \frac{P(X|\theta_j)T(\theta_j)}{P(X)}, T(\theta_j) = 1/\max[P(X|\theta_j)/P(X)] \quad (3)$$

$T(\theta_j)$ in (3) may be called longitudinally normalizing constant, which makes the maxima of $T(\theta_j|x)$ be 1.

Note that the statistical probability is normalized whereas the logical probability is not. In general, there are $P(y_0)+P(y_1)+\ldots+P(y_n)=1$ and $T(\theta_0)+T(\theta_1)+\ldots+T(\theta_n)>1$. For example, when $V=\{y_0=\text{"Non-adult"}, y_1=\text{"Adult"}, y_2=\text{"Youth"}, y_3=\text{"The Old"}\}$, $T(\theta_0)+T(\theta_1)=1$, $T(\theta_0)+T(\theta_1)+T(\theta_2)+\ldots>1$.

*2.3 Relationship between Likelihood and Cross-entropy*

Definition 2 : $\mathbf{X}_j$ is a sub-sample or sequence of data points $x(1), x(2), \ldots, x(N_j) \in U$ with label $y_j$. $\mathbf{D}$ is a sample or sequence of examples $\{(x(t), y(t))|t=1 \text{ to } N; x(t)\in U; y(t)\in V\}$, which includes $n$ different sub-samples $\mathbf{X}_j$, $j=1, 2, \ldots$ If $\mathbf{D}$ is large enough, we can obtain distribution $P(x, y)$ from $\mathbf{D}$, and conditional distribution $P(x|y_j)$ from $\mathbf{X}_j$.

When the data points in $\mathbf{X}_j$ come from Independent and Identically Distributed (IID) random variables, we have

$$\log P(x(1), x(2), \ldots, x(N_j)|\theta_j) = \log \prod_i P(x_i|\theta_j)^{N_{ji}}$$
$$= N_j \sum_i P(x_i|y_j)\log P(x_i|\theta_j) = -N_j H(X|\theta_j). \quad (4)$$

where $H(X|\theta_j)$ is a cross-entropy. Hence, the ML criterion is equivalent to the least cross-entropy criterion or the Least Kullback-Leibler divergence criterion [2].

*2.4 Semantic Information Measure*

The (amount of) semantic information conveyed by $y_j$ about $x_i$ is defined with log-normalized-likelihood [12, 14]:

$$I(x_i; \theta_j) = \log \frac{P(x_i|\theta_j)}{P(x_i)} = \log \frac{T(\theta_j|x_i)}{T(\theta_j)} \quad (5)$$

It means that the larger the deviation is, the less information there is; the less the logical probability is, the more information there is; and, a wrong hypothesis $y_j$ may convey negative information — these conclusions accord with Popper's thought [22]. If $T(\theta_j|x)\equiv 1$, $I(x_i; \theta_j)$ will becomes Bar-Hillel-Carnap's semantic information measure [23].

To average $I(x_i; \theta_j)$, we have

$$I(X;\theta_j) = \sum_i P(x_i|y_j)\log\frac{P(x_i|\theta_j)}{P(x_i)} = \sum_i P(x_i|y_j)\log\frac{T(\theta_j|x_i)}{T(\theta_j)} \quad (6)$$

where $P(x_i|y_j)$ ($i=1,2,\ldots$) is the sampling distribution, which may be unsmooth or discontinuous. This formula is like Donsker-Varadhan representation in [6, 24].

To average $I(x_i; \theta_j)$ in Eq. (5) for different $X$ and $Y$, we have the Semantic Mutual Information (SMI)

$$I(X; \theta) = \sum_j \sum_i P(x_i, y_j)\log\frac{P(x_i|\theta_j)}{P(x_i)}$$
$$= \sum_j \sum_i P(x_i, y_j)\log\frac{T(\theta_j|x_i)}{T(\theta_j)} \quad (7)$$

For an unbiased estimation $\theta_j$, the truth function or membership function may be expressed by a Gaussian distribution without coefficient: $T(\theta_j|X)=\exp[-(X-\mu_j)^2/(2\sigma^2)]$. Hence

$$I(X; \theta) = H(\theta) - H(\theta|X) =$$
$$-\sum_j P(y_j)\log T(\theta_j) - \sum_j \sum_i P(x_i, y_j)(X-\mu_j)^2/2\sigma^2 \quad (8)$$

If we replace $-(X-\mu_j)^2/(2\sigma^2)$ with $T_\theta$, then $I(X; \theta)$ will become the neural information measure defined in [6].

It is easy to find that the maximum SMI criterion is a special Regularized Least Squares (RLS) criterion [21]. $H(\theta|X)$ is the mean squared error, and $H(\theta)$ is the negative regularization term. It is different from others who use $I(X;Y)$ as regularization term [11] that we use $I(X; Y)$ as special RLS.

Lu [12, 14] extended the rate distortion function $R(D)$ to obtain the $R(G)$ function where $G$ is the lower limit of $I(X; \theta)$, and $R$ is the minimum Shannon's mutual information. $R(G)$ is bowl-like and has a matching point where $R(G)=G$. Apart from this point, $R(G)>G$. $R(G)$ function can be used to prove the convergence of the MMI classification of unseen instances [3] and the convergence of the CM-EM algorithm for mixture models [26].

# 3 The Classifications of Unseen Instances

*3.1 The CM Algorithm*

For the MMI classification of visible instances, Lu has used the CM algorithm for multi-label classifications [26]. In these cases. we only need multi-label learning and multi-label classification one time. The multi-label learning is to obtain a semantic channel that consists of a group membership functions $T(\theta_j|x), j=0,1,…$ by the third kind of Bayes' theorem for continuous Transition Probability Functions (TPF) [1] $P(y_j|X), j=0,1,…$ or by Eq. (6) for discontinuous TPFs.

For the classification, we can use such classifier:

$$y_j *= h(X) = \arg\max_{y_j} \log I(\theta_j; x_i)$$
$$= \arg\max_{y_j} \log \frac{T(\theta_j|X)}{T(\theta_j)} \quad (9)$$

However, for unseen instances, we need $Y=f(Z)$ instead of $Y=h(X)$ and an iterative method.

We use the medical test as the example shown in Fig.1 to explain the relationships between $X, Y,$ and $Z$ in unseen instance classifications.

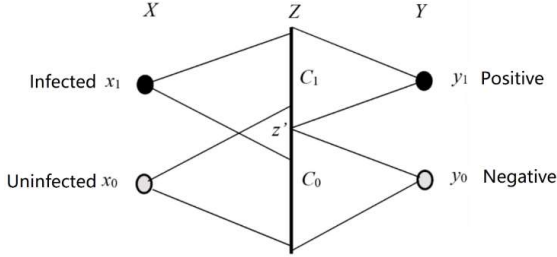

Fig. 1 Relationships between $X, Y,$ and $Z$ in Medical tests. The $x$ is a true class or true label. The $y$ is a selected label. The $z'$ is the dividing point.

Let $C_j$ be a subset of $C$ and $y_j=f(z|z \in C_j)$. Hence $S=\{C_1, C_2, …\}$ is a partition of $C$. Our aim is, for given $P(x, z)$ from **D**, to find optimized $S$, which is

$$S^* = \arg\max_{S} I(X; \theta|S)$$
$$= \arg\max_{S} \sum_j \sum_i P(C_j)P(x_i|C_j)\log \frac{T(\theta_j|x_i)}{T(\theta_j)} \quad (10)$$

First, we obtain the Shannon channel for given $S$:

$$P(y_j|x) = \sum_{z_k \in C_j} P(z_k|x), j=0,1,... \quad (11)$$

From this Shannon's channel, we can obtain the semantic channel $T(\theta|X)$ in numerical values (Matching I) [3]. Then, for given $Z$ we have conditional information:

$$I(X;\theta_j|Z) = \sum_i P(x_i|Z) \log \frac{T(\theta_j|x_i)}{T(\theta_j)}, j=0,1,... \quad (12)$$

which are some curved surfaces over the feature space. We may also use $P(y_j|x)/P(y_j)$ instead of $T(\theta_j|x)/T(\theta_j)$. Then we let the Shannon channel match the semantic channel by the classifier (Matching II)

$$y_j *= f(Z) = \arg\max_{y_j} I(X;\theta_j|Z), j=0,1,... \quad (13)$$

Repeat (11)-(13) until $S$ does not change. The convergent $S$ is $S^*$ we seek. Some iterative examples show that the above algorithm is fast and reliable. The convergence can be proved with the help of the $R(G)$ function [3].

*3.2 A Two-dimensional Example with Different Initial Partitions to Test the CM Algorithm*

From a sample $\mathbf{D}=\{(x(t), z(t))|t=1,2,…, N; x(t) \in U, z(t) \in C)\}$, we can obtain $P(Z|x_i), i=0, 1, 2$. Then, we repeat Eqs. (11)-(13) to achieve the MMI classifications. If $P(Z|x_i)$ is not smooth, we may replace it with

$$P^*(Z|\theta_{xi}) = \underset{\theta_{xi}}{\mathrm{argmax}}\, I(Z;\theta_{xi})$$
$$= \underset{\theta_{xi}}{\mathrm{argmax}} \sum_i P(z_k|x_i) \log \frac{P(z_k|\theta_{xi})}{P(z_k)} \quad (14)$$

We have tested the CM algorithm by some examples.

**Example 1**: The $z$ changes from 0 to 100. Two Gaussian distributions $P(z|x_0)$ and $P(z|x_1)$ have parameters $\mu_0=30, \mu_1=70, \sigma_0=15,$ and $\sigma_1=10$. $P(x_0)=0.8$ and $P(x_1=0.2)$. The start point $z'$ is 50.

**The iterative process**: The first iteration makes $z'=53$. The second iteration makes $z'=54$. The third iteration makes $z^*=54$.

**Example 2:** The $z$ is two-dimensional. There are three true classes $x_0, x_1,$ and $x_2$ with probability distribution $P(x)$ as shown in Fig 4. $P(Z|x_0)$ (red) and $P(Z|x_1)$ (green) are two Gaussian distributions, and $P(Z|x_2)$ (blue) is the mixture of two Gaussian distributions $P(Z|x_{21})$ and $P(Z|x_{22})$. About mixture models, Lu proposed an improved EM algorithm, the CM-EM algorithm, for fast and reliable convergence [28]. The parameters of four Gaussian distributions are listed in Table 1.

Table 1. The parameters of four Gaussian distributions

|              | $\mu_m$* | $\mu_n$ | $\sigma_m$ | $\sigma_n$ | $\rho$ | $P(x_i)$ |
|---|---|---|---|---|---|---|
| $P(Z|x_0)$   | 50  | 50  | 75  | 200 | 50  | 0.2 |
| $P(Z|x_1)$   | 75  | 90  | 200 | 75  | -50 | 0.5 |
| $P(Z|x_{21})$| 100 | 50  | 125 | 125 | 75  | 0.2 |
| $P(Z|x_{22})$| 120 | 80  | 75  | 125 | 0   | 0.1 |

*$m$ and $n$ are horizontal and vertical coordinates of $Z$.

**The iterative process**: The initial partition is made by two vertical lines. After two iterations, the partition is expressed by three curves as shown in Fig. 4.

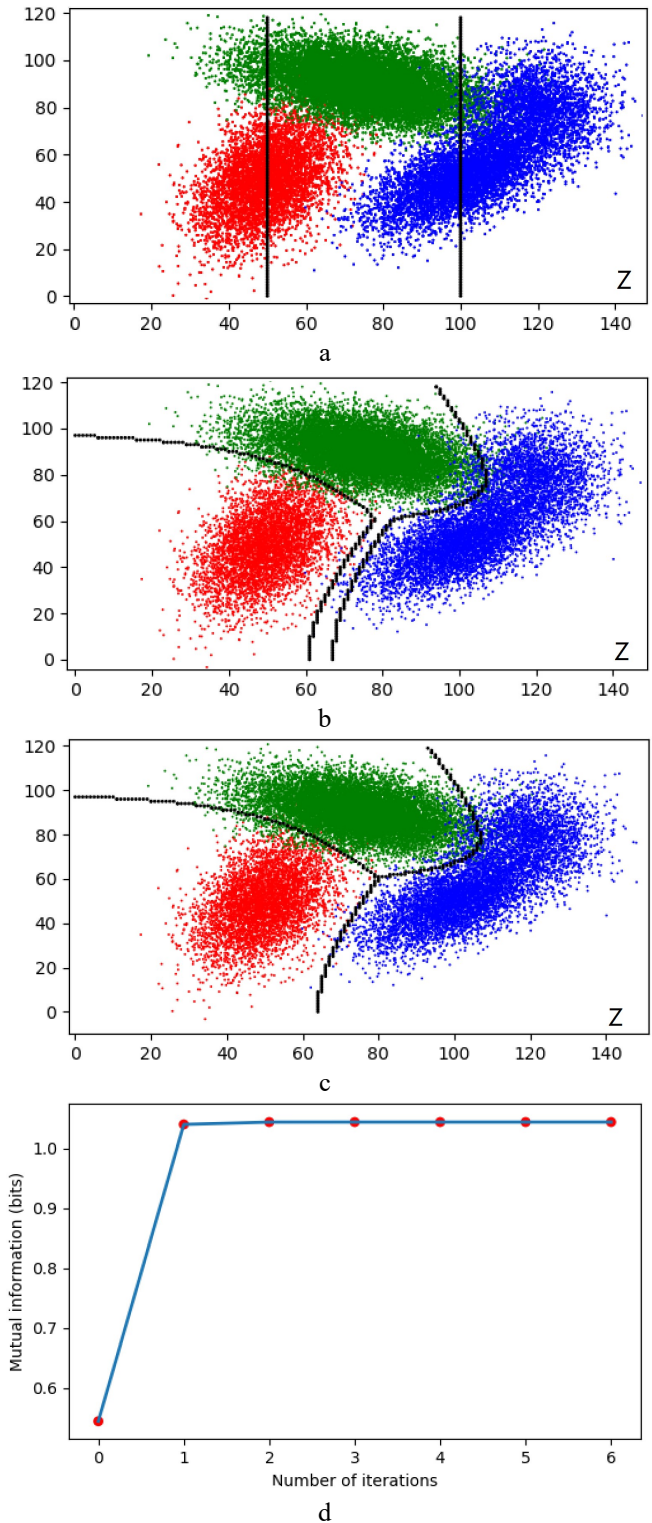

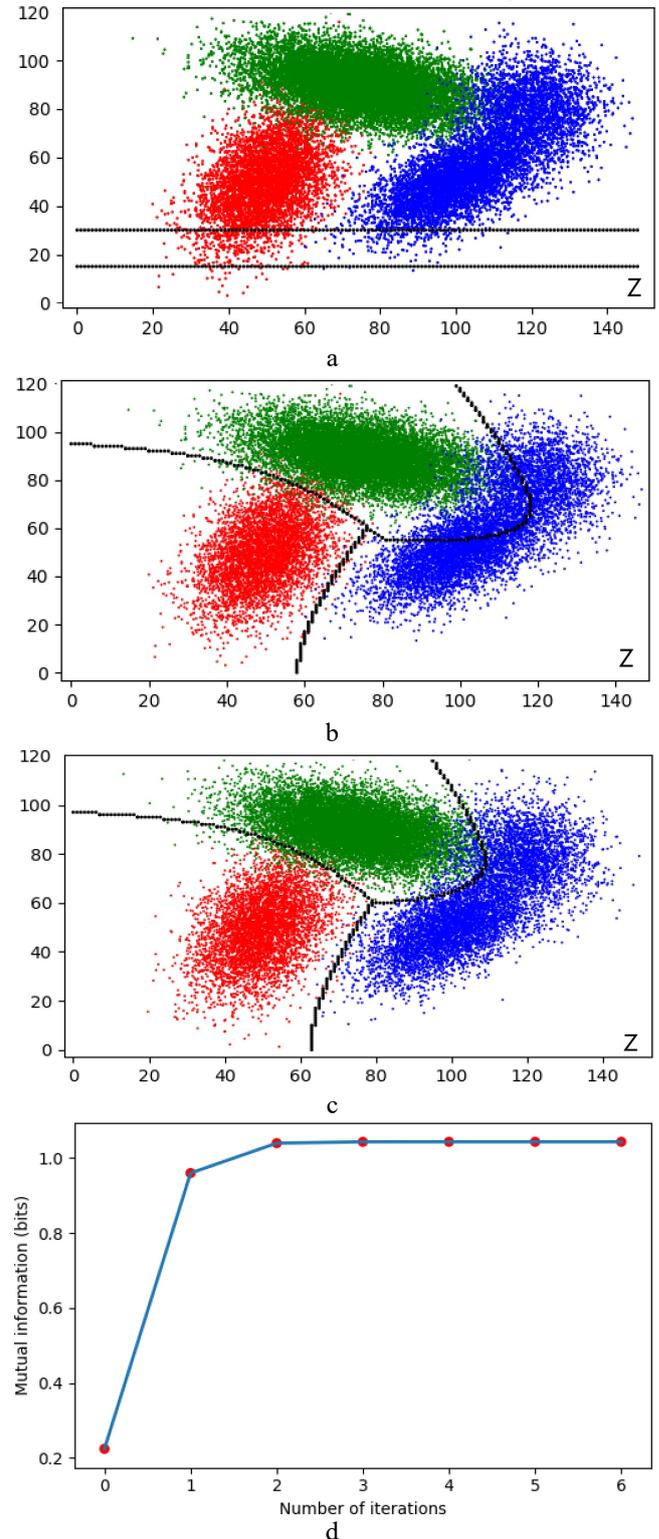

Fig. 2 The iterative experiment with a simple initial partition.
(a) The initial partition with two vertical lines
(b) The partition after the first iteration
(c) The partition after the second iteration
(d) The Mutual Information (MI) changes with iterations. After two iterations, the MI is 1.0434 bits, which reaches 99.99% of the MMI 1.0435 bits.

Fig. 2 (d) shows That the convergence is very fast. To test the reliability of the CM algorithm, we use a very bad initial partition by two horizontal lines as shown in Fig. 3 (a). The iterative process is shown in Fig. 3.

Fig. 3 The iterative experiment with a very bad initial partition. After two iterations, the MI is 1.0397 bits, which reaches 99.66% of the MMI 1.0435 bits.

# 4 Discussions

*4.1 Distinctions between the MMI Criterion and the LER Criterion*

Hu [9] has discussed the distinctions of two criterions. Now we further discuss their characteristics for unseen instance classifications.

Since SMI is average log normalized likelihood and its upper limit is Shannon's mutual information, the MMI criterion is compatible with the ML criterion. If we use $P(x|\theta_j)$, $P(x|\theta_j)/P(x)$, or $P(\theta_j|x)/P(\theta_j)$ as the likelihood function, the ML estimations and classifications are the same. If we use $P(x_i|Z, \theta)$, e. g., $P(\theta_{xi}|Z)$, as the likelihood function and use the Bayes classifier

$$y_i^* = f(Z) = \arg \max_i P(\theta_{xi}|Z) \quad (15)$$

then this classifying criterion is the Maximum Posterior Probability (MPP) criterion in fact, which is equivalent to the LER criterion (see Fig. 4 for details).

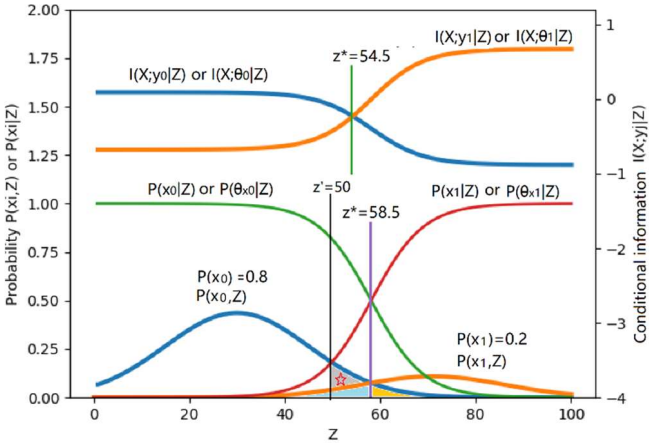

Fig. 4 In comparison with the MPP or LER criterion, the MMI criterion can reduce failing to report smaller probability events. After $P(x_0)$ is changed from 0.5 to 0.8, if we still use z'=50, the error rate will increase a ratio indicated by the area with a star.

Our Experiments shows that when $P(X)$ is uniform, the MMI criterion is equivalent to the MPP or LER criterion whereas when $P(X)$ is non-uniform, they are different. For example, in the medical test as shown in Fig.1, where $Z$ changes from 0 to 100, $P(Z|x_0)$ and $P(Z|x_1)$ are two Gaussian distributions. If $P(x_0)=P(x_1)=0.5$, both criterions make $z^*$ between 50 and 51. If $P(x_0)=0.8$ and $P(x_1)=0.2$, then $z^*$ will locate between 54 and 55 with the MMI criterion and between 58 and 59 with the MPP or LER criterion. The reason is that the MMI criterion can reduce failing to report small probability events.

Some people have obtained classifying results with the ML or MMI criterion. Then they still use the LER or maximum correctness rate criterion to evaluate the results or compare the used algorithm with other algorithms. We think that this is unnecessary or unfair.

*4.2 Comparing the CM Algorithm and the Gradient Descent for Low-Dimensional Feature Space*

The MMI classification introduced in Reference [8] is only for visible instances. The MMI classification of unseen instances are often used for maximizing average (for different $y$) log likelihood, where the optimized partition $S^*$ is also needed. The MMI estimation [6] also needs or results in the MMI classification of unseen instances.

Popular methods for the MMI classifications use parameters to construct the partitioning boundaries between classes and then optimize these parameters by the gradient descent or Newton's method. However, the above CM algorithm separately constructs $n$ likelihood functions by parameters for $n$ different classes and then labels every $Z$ or provides the numerical solution of boundary partition.

Table 2 compares the CM algorithm and the gradient descent.

Table 2 Comparison of the CM algorithm and the gradient descent for low-dimensional feature spaces

| About | Gradient descent | CM algorithm |
| --- | --- | --- |
| Models and parameters | For boundary partition | For every class |
| Boundaries presented by | Functions with parameters | Numerical values |
| Gradient and search | necessary | unnecessary |
| Convergence | Not easy | Easy |
| Computation | Complicated | Not complicated |
| Num. of iterations | Many | 2-3 |
| Samples required | Not necessarily big | Big enough |

The CM algorithm requires that every sub-sample for every class is big enough so that we can construct $n$ likelihood functions for $n$ classes.

*4.3 Combining the CM algorithm and Neural Networks for High-dimensional Feature Spaces*

If $Z$ is high-dimensional, it is infeasible to label every possible $Z$. In this case, we can use neural networks to partition feature spaces. A neural network is a classifier $Y=f(Z)$. For a given neural network, e. g., a partition, Matching I is to let the semantic channel match the Shannon channel to obtain reward functions $I(X; \theta_j |Z)$, $j=0, 1,...$ For given reward functions, Matching II is to let the Shannon channel match the semantic channel to obtain new network parameters. Repeating two steps can make $I(X; \theta)$ converge to the MMI. The Matching I and Matching II are like the tasks of generative and discriminative models in GAN [27]. The CM algorithm is different from MINE [6] in that we needn't use parameters to construct $I(x_i; \theta_j)$ or $\log(e^T/Z)$ in [5], which may be equal to $I(x_i; y_j)=\log[P(y_j|x_i)/P(y_j)]$ according to the last partition. We only need parameters for the classifier $Y=f(Z)$. The convergence can be proved in the same way [3]. Combining the CM algorithm and existing deep learning methods [28], we should be able to speed the MMI classifications for high-dimensional feature spaces.

## 5. Conclusions

Using the CM algorithm for the MMI classifications of unseen instances, when feature spaces are low-dimensional, we may use numerical values instead of models with parameters to express partitioning boundaries. Without needing gradient and search, this algorithm is very simple and fast. We provided a typical example where the feature space is two-dimensional. To show the efficiency and reliability of the algorithm, we tested the algorithm by different initial partitions. With a simple initial partition, after two iterations, the MI reaches 99.99 of the maxima. With a very bad initial partition, after two iterations, the MI reaches 99.66 of the maxima. We compared the CM algorithm with the popular gradient descent. The CM algorithm is faster and more reliable for low-dimensional feature spaces. This algorithm also has a disadvantage that it requires $n$ big enough samples for $n$ classes. For high-feature spaces, we may combine the CM algorithm with neural networks to achieve the MMI classifications. For this purpose, we need further studies.

Lu's Homepage: http://www.survivor99.com/LCG/english/
Lu's more studies about statistical learning: http://www.survivor99.com/LCG/CM/Recent.html